\documentclass[11pt]{article}

\usepackage[preprint]{acl}

\usepackage{times}
\usepackage{latexsym}

\usepackage[T1]{fontenc}

\usepackage[utf8]{inputenc}

\usepackage{microtype}

\usepackage{inconsolata}

\usepackage{graphicx}

\usepackage{amsmath}
\usepackage[percent]{overpic}
\usepackage{placeins}
\usepackage{url}

\title{Self-Supervised Speech Representations Track Spoken Language Convergence to Adult Models in Infants and Children Who Are Deaf/Hard-of-Hearing}

\author{
  Landon Choy, Ali Sartaz Khan, Sonia Patrizi, \\
  \textbf{Daisy Ye, Julianna Gross, Margaret Cychosz} \\
  Stanford University, USA\\
  \texttt{mcychosz@stanford.edu}
}

\begin{document}
\maketitle
\begin{abstract}
Language development is characterized by a gradual convergence of children's speech toward adult patterns. Measuring this process has traditionally required detailed transcription and language-specific expertise, limiting scalability across languages and populations. Here, we use speech embeddings to capture this convergence directly from the acoustic signal in longform, child-centered recordings, taken as children go about their daily lives. Using HuBERT-BASE, we extracted embeddings from speech vocalizations of children who are deaf/hard-of-hearing and their female adult caregivers ($>$925 hrs. observation). Embedding distance between children and caregivers decreased with hearing age, controlling for pitch and vocalization length, indicating, as expected, that children's speech patterns converge to caregivers over development. This single distance metric likewise related to multiple standardized measures of speech and language from infancy through preschoolhood. These results suggest a path toward scalable, language-neutral assessment of spoken language development from children's everyday lives. 

\end{abstract}

\section{Introduction}

Automatic recognition of children's spontaneous speech offers an opportunity to derive objective measures of language development from audio, forgoing the costly transcription and linguistic expertise traditionally required \citep{demuth2006providence}. However, naturalistic, usually child-centered, recordings contain background noise, overlapping speech, and are plagued by diarization errors, making it difficult to derive meaningful language signals or developmental metrics \citep{li2025challenges, peurey2025challenges2}. Developing methods that remain robust under these conditions is an open problem for automatic speech recognition research.

Recent work has begun to address this challenge. \citet{sy2023entropymetric} proposed an unsupervised metric of language development based on the entropy of discretized speech units derived from HuBERT-BASE embeddings \citep{hsu2021hubert}. Entropy quantified how surprising children’s speech was under an adult-trained language model, with lower entropy indicating increasing convergence toward adult speech. While this approach failed under noisy naturalistic recordings, experiments with clean synthetic speech recovered the expected pattern of child entropy converging toward caregivers. \citet{ott2026cpdev} showed that deriving canonical proportion, the proportion of well-formed consonant-vowel transitions, from naturalistic, child-centered audio predicts preschoolers' performance on numerous standardized measures of speech and language. But this measure is somewhat coarse, and may not capture finer-grained developments in children's speech or language development. 


In this paper, we propose a simple framework that models language development as the continuous distance from caregiver speech in embedding space, without requiring transcription or discrete linguistic units. We apply this framework to children who are deaf or hard-of-hearing (DHH), who are at increased risk for language delay because their reduced access to spoken language can affect early language acquisition, particularly prior to intervention. Following intervention (e.g., hearing aids or cochlear implants), these children are under clinical care to ensure developmental progress, yet existing speech-language assessments are difficult to administer frequently and reliably, especially in infants and toddlers. This gap motivates scalable measures that can model speech-language development directly from children’s everyday speech. The simplicity of our approach makes it computationally lightweight and practical for longitudinal follow-up of post-intervention progress. 

We make the following contributions: 1) introduce an embedding-space distance metric computed directly from raw audio representations, capturing fine-grained developmental changes in children's speech-language structure relative to adult models; 2) show that this metric significantly predicts  evaluated speech and language outcomes in children aged 10--65 months; and 3) demonstrate robustness to diarization errors through bootstrapping and sensitivity analyses, yielding stable estimates under input perturbations.

\section{Methods}

Participants were 34 children with bilateral moderate-profound hearing loss (16f/18m; 27 bilateral cochlear implant, 3 bimodal cochlear implant+hearing aid, 2 bilateral hearing aid, 2 unilateral cochlear implant); see Table~\ref{tab:descriptives} for age detail. 32/34 children were exposed to English >50\%; two children were also exposed to some Mandarin (N=1) or Spanish (N=1). N=14 children (41\%) contributed data from two or more longitudinal timepoints (M=2.8, SD=1.1, total observations in the dataset=59). Data collection was approved by the Institutional Review Boards at the authors' institution at the time of data collection. 

Each child wore a Language ENvironment Analysis (LENA) recording device in a specialized shirt, capturing both surrounding speech and the child’s own vocalizations. Families were instructed to activate the recorder once the child awoke and to record for up to 16 hours. Each recording, corresponding to a single child-timepoint, averaged 15.7 hours in length (SD = 1.3, range = 9.2–16), yielding a dataset of 925 hours of child-centered audio. 

\subsection{Speech-language assessments}

Children/caregivers additionally completed a number of standardized speech-language assessments. \textit{Parent-reported vocabulary}: Parents of children aged $\le 52$ mos. at study onset\footnote{Children who are DHH often have language delay; clinical judgment was applied to continue employing this assessment beyond the typical age range.} completed the American English MacArthur-Bates Communicative Development Inventory (MB-CDI) \citep{fenson2007mbcdi}, a checklist of words the child knows and understands. Because we had recordings of the child's speech production in the long-form recordings, and not, for example, speech comprehension in a controlled task, here we model the number of words that children produced as one of our outcome measures in the results. \textit{Child vocabulary} was measured in children $\ge 37$ mos. at study onset with the Peabody Picture Vocabulary Test-4 (PPVT-4) \citep{dunn2007ppvt4}, which indexes children's receptive vocabulary size, and the Expressive Vocabulary Test-2 (EVT-2) \citep{williams2007evt2}, which indexes expressive vocabulary size. \textit{Child speech articulation} was assessed in children $\ge 37$ mos. using the Goldman-Fristoe Test of Articulation-2 (GFTA-2) \citep{goldman2000gfta2}, where consonant articulation accuracy is assessed across word positions (e.g. initial, medial) in a picture-naming task. Responses were scored offline by two trained research assistants. Speech-language assessments were conducted within $60$ days (M=19.3, SD=22.0) of the child's longform recording to assess concurrent relationships between the spontaneous speech collected in the longform naturalistic recording and controlled speech-language patterns. 

\subsection{Processing pipeline}

\begin{figure*}
  \centering
  \includegraphics[width=0.75\linewidth]{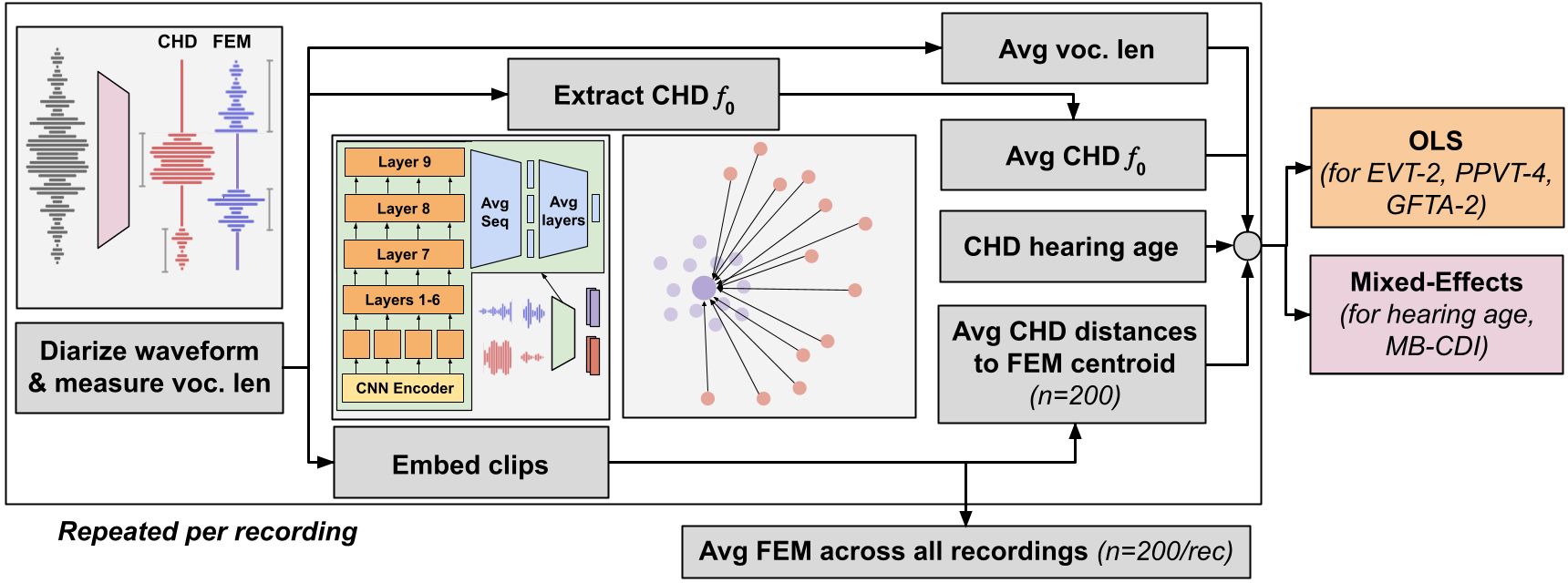}
  \caption{Processing pipeline overview. Longform recordings are speaker diarized (child `CHD', adult female `FEM') and embedded using HuBERT-BASE \citep{hsu2021hubert}. Audio was sampled at 16\,kHz with a 20\,ms frame rate, producing 768-dimensional vectors, which were extracted from layers 7--9 and mean-pooled across time and layers. Vocalization lengths were computed as the durations of speaker-diarized speech segments. Fundamental frequency ($f_0$) was extracted from subsampled key child (CHD) clips. Female adult (FEM) embeddings, representing adult female caregivers (e.g., mothers) from all recordings, were subsampled (small blue points) to create a global caregiver centroid (large blue point; visualized here as a 2D projection of the 768-dimensional embedding space). CHD embedding distance to this centroid was measured for each recording (red points) and used in statistical models as one observation/timepoint alongside mean $f_0$, mean vocalization length, and hearing age.}
  \label{fig:workflow}
\end{figure*}

\textit{Recording speaker diarization} was conducted by segmenting the continuous audio stream into speaker vocalizations using the LENA interpreted time segment `.its' (a proprietary data file format associated with each LENA recording) speaker labels. Audio segments classified as the key child (CHD; M=3530, SD=1346/rec) and adult female near the child (FEM; M=2391, SD=1141/rec) were retained (see Fig~\ref{fig:workflow} for the full processing pipeline).

\textit{Embeddings} were extracted from HuBERT-BASE, chosen because of its self-supervised pretraining mechanism that transfers well for downstream speech tasks across languages for both adult and child speech  \citep{boito2024mhubert147, charlot2026babyhubert}. Embeddings were extracted from layers 7–9 following \citet{charlot2026babyhubert}, who used layer 7 to generate pretraining clustering targets for downstream child voice type classification. Since the clustering procedure identifies hidden units corresponding to acoustic speech units, \citet{charlot2026babyhubert}'s pretraining procedure suggests HuBERT’s mid-upper layers encode salient acoustic structure.

\textit{Fundamental frequency} ($f_0$) was extracted using the PYIN algorithm \citep{mauch2014pyin}. Here, $f_0$ is a control because speech representations encode $f_0$-related variation, particularly in lower transformer layers \citep{lin2022sslpitchlayers}. Since $f_0$ decreases with anatomical growth over development, it could potentially confound linguistic convergence with acoustic maturation \citep{Lee1999agepitchcorr}. Thus, we included each child’s per-recording mean $f_0$ as a covariate to isolate linguistic from non-linguistic variance in embedding distance.

\textit{Vocalization length} was extracted from the .its files and likewise included as a covariate to avoid a potential confound with language development as children's speech utterances lengthen with age \citep{rice2010mean, ramsdell2018utterance}.

\begin{table}
  \centering
  \small
  \begin{tabular}{l l l}
    \hline
    \textbf{Variable} & \textbf{N$_\text{obs}$/N$_\text{child}$} & \textbf{M (SD)} \\
    \hline
    \multicolumn{3}{l}{\textit{Recording details}} \\
    \quad Chron.\ age (mos) & 59/34 & 36.2 (14.4) \\
    \quad Hearing age (mos) & 59/34 & 22.2 (14.3) \\
    \quad Key child vocs/rec       & 59/34 & 3531 (1347) \\
    \quad Adult female vocs/rec       & 59/34 & 2392 (1142) \\
    \quad Duration (s)      & 59/34 & 16.1 (3.8) \\
    \hline
    \multicolumn{3}{l}{\textit{Language outcomes}} \\
    \quad MB-CDI            & 36/16 & 107.0 (166.3) \\
    \quad PPVT-4            & 22/17 & 95.6 (19.4) \\
    \quad EVT-2             & 23/18 & 100.1 (17.6) \\
    \quad GFTA-2            & 22/17 & 72.4 (17.3) \\
    \hline
  \end{tabular}
  \caption{Descriptive stats. N$_\text{obs}$=recording--assessment pairs; N$_\text{child}$=unique children; tp = timepoint. Hearing age=time since implantation/amplification. MB-CDI=raw words produced; PPVT-4, EVT-2, GFTA-2=standard scores.}
  \label{tab:descriptives}
\end{table}

\textit{Embedding distance} was computed by constructing a global FEM centroid across \textit{all} recordings, from all adult females, and comparing individual CHD vocalizations from each recording to this reference. Distances were mean-pooled at the child–timepoint level to yield a single observation per pair (Fig.~\ref{fig:workflow}), to model each child’s speech relative to a single, stable adult reference.

\subsection{Sensitivity analysis}

Speaker misattribution is a major concern in longform child-centered recordings. Following \citet{gautheron2026errors}, we conducted sensitivity analyses to simulate diarization errors and test whether the observed relationships could arise from speaker-label contamination. Specifically, increasing proportions of CHD vocalizations ($k$\%) were randomly replaced with vocalizations from other speaker classes ($k=0$–100\% in 10\% increments). Embedding distances and downstream models were recomputed at each contamination level. Full details are provided in App.~\ref{app:sens_analysis_methods}.

\section{Results}
\label{sec:results}

We begin by characterizing the embedding distance metric, and then evaluate its relationship with the children’s hearing age to establish how it behaves given known language development trajectories. We then assess the relationship between embedding distance and standardized speech-language measures to test whether greater convergence to adult speech indexes more mature speech and language skills\footnote{Code for all analyses is available at: \url{https://github.com/spoglab-stanford/cld-indexing}}.

When the longitudinal structure of the data permitted (i.e. multiple observations per child), we fit linear mixed-effects models with a random intercept for child using \texttt{mixedlm} from \texttt{statsmodels}; otherwise, we fit ordinary least squares models using \texttt{ols} \citep{seabold2010statsmodels}. Mean CHD $f_0$ from each recording was included as a covariate in all models to control for pitch-related variance in the embedding space (see Methods for detail). Hearing age and vocalization length were also included to account for residual developmental variance not captured by $f_0$. Continuous predictors were z-score normalized. Model fit was evaluated using AIC and likelihood-ratio tests comparing models with and without the embedding distance term, alongside fixed-effect coefficient significance. We also report incremental variance explained ($\Delta R^2$), the increase in $R^2$ attributable to each predictor when added to the model.

For each recording, 200 CHD vocalization embeddings/recording were sampled (from up to 800 observations) to compute the average distance to the FEM centroid. This subsampling was repeated for 1000 iterations, refitting the statistical models each time to obtain $95\%$ bootstrap confidence intervals over the estimated coefficients and model metrics. The FEM centroid was fixed across iterations and computed from caregiver vocalizations pooled across all child-timepoints ($n=59$), totaling 11,800 FEM vocalizations (200 per timepoint). The bootstrap intervals therefore capture variability in the sampled CHD observations and provide robustness to diarization errors.

\begin{figure*}[t]
\centering
\begin{overpic}[width=0.8\linewidth]{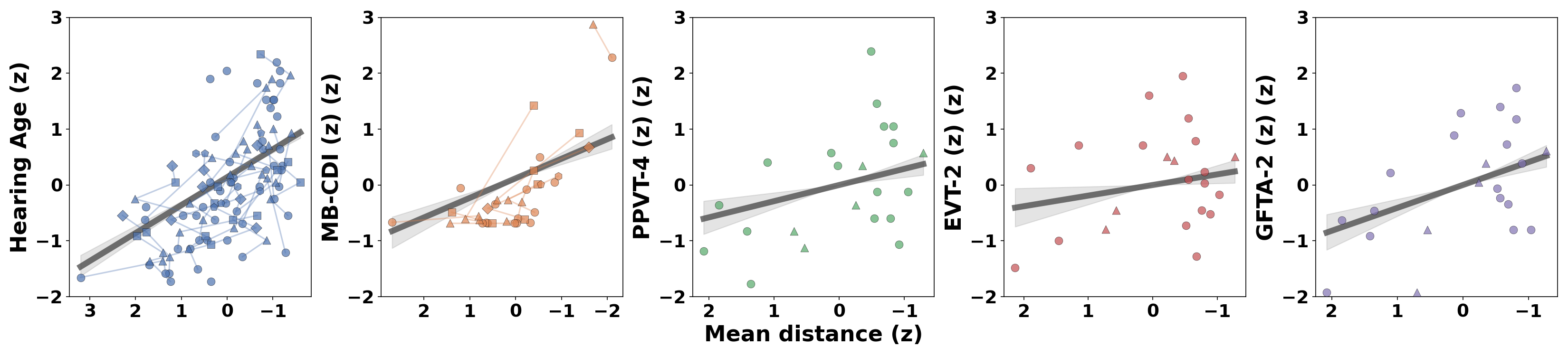}
\put(1,0){(a)}
\put(21,0){(b)}
\put(41,0){(c)}
\put(61,0){(d)}
\put(81,0){(e)}
\end{overpic}
\caption{Relationships between CHD--FEM embedding distance and developmental measures. Points show child--timepoint mean distance from CHD embeddings to the FEM centroid, averaged across sampled clips (and bootstrap resamples). Marker shape indicates within-child timepoint order: first = circle, second = triangle, third = square, fourth = diamond, fifth = pentagon, and sixth = hexagon. Lines connect successive observations for the same child; connecting lines  omitted for PPVT-4, EVT-2, and GFTA-2 due to limited repeated observations per child. The distance axis is reversed so that reduced CHD--FEM distance appears to the right, such that all panels show a negative trend. The gray line shows the fixed-effect relationship between distance and the outcome, computed from the mean bootstrap model coefficients; shaded ribbon indicates the pointwise 95\% confidence interval obtained from the distribution of bootstrap-fitted lines.}

\label{fig:model_fits}
\end{figure*}
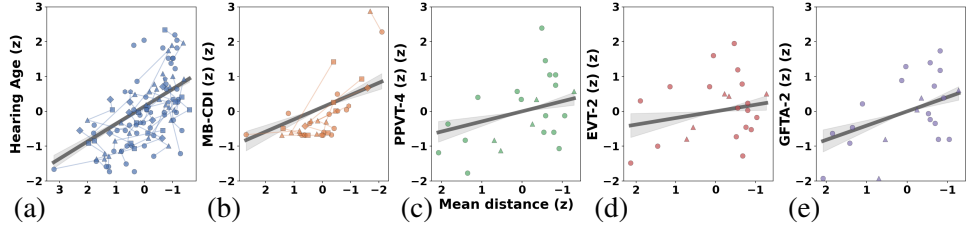

\begin{table*}[!t]
  \centering
  \small
  \begin{tabular}{l c c c c c}
    \hline
    \textbf{Statistic} & \textbf{Hearing Age} & \textbf{MB-CDI} & \textbf{PPVT-4$^\dagger$} & \textbf{EVT-2$^\dagger$} & \textbf{GFTA-2$^\dagger$} \\
    \hline

    \multicolumn{6}{l}{\textit{Fixed-effect coefficients (with distance)}} \\

    \quad Distance ($\beta$)     
        & -0.50***
        & -0.35*** 
        & -0.29*** 
        & -0.19* 
        & -0.41*** \\
    
    \quad Hearing age ($\beta$)  
        & 
        & 0.53*** 
        & 0.45*** 
        & 0.34*** 
        & 0.34*** \\
    
    \quad Mean f0 ($\beta$)      
        & -0.01
        & -0.09
        & -0.28*** 
        & -0.34*** 
        & -0.10 \\
    
    \quad Mean voc. len ($\beta$)      
        & 0.01
        & 0.12*** 
        & 0.08
        & 0.08
        & <0.001 \\
    
    \hline
    \multicolumn{6}{l}{\textit{Model comparison (with vs. without distance)}} \\
    
    \quad $\Delta$AIC            
        & -24.68***
        & -10.28*** 
        & -0.84 
        & 0.79
        & -2.74 \\
    
    \quad LR test $\chi^2(1)$
        & 26.68***
        & 12.28***
        & 2.84***
        & 1.21*
        & 4.74*** \\
    
    \quad LR $p$            
        & <0.001***
        & <0.001*** 
        & 0.13 
        & 0.35 
        & 0.05 \\
    
    \hline
    \multicolumn{6}{l}{\textit{Incremental variance explained ($\Delta R^2$; beyond baseline predictors)}} \\
    
    \quad Distance (\%)          
        & 11.30***
        & 9.47*** 
        & 6.95*** 
        & 3.51** 
        & 14.33*** \\
    
    \quad Hearing age (\%)       
        & 
        & 11.11*** 
        & 18.22*** 
        & 10.38*** 
        & 10.39*** \\
    
    \hline
    \end{tabular}
    \caption{Bootstrap model summaries. Coefficients correspond to the full model (with distance metric). $\Delta$AIC and LR test compare models with vs. without distance, such that negative $\Delta$AIC values favor the inclusion of our metric. ($\Delta R^2$)=drop-one variance change in model $R^2$ attributable to each predictor beyond the others. Values show bootstrap means across CHD resamples; significance markers denote bootstrap significance (* $p<0.05$, ** $p<0.01$, *** $p<0.001$). Bootstrap inference reflects CHD resampling, whereas LR test $\chi^2$ statistics reflect model-level comparisons. $^\dagger$=OLS; others=mixed-effects with random intercept/child.}
    \label{tab:ci_stats}
\end{table*}
\subsection{Does embedding distance decrease with increased hearing experience?}

We first tested the relationship between embedding distance and children's hearing age, or their experience with spoken language. We expect a gradual convergence of child to adult speech in the embedding space as children gain more experience. We found embedding distance improved baseline model fit and significantly predicted hearing age ($\beta=-0.50$, $p<.001$; Table~\ref{tab:ci_stats}; Fig~\ref{fig:model_fits}a). Model comparisons further supported including embedding distance ($\Delta$AIC$=-24.68$, $p<.001$), and it accounted for substantial additional variance ($11.30\%$, $p<.001$). This indicates that hearing age, reflecting cumulative experience with spoken language, is associated with reduced child-adult embedding distance, suggesting progressive convergence toward adult speech patterns with increasing auditory experience. Hearing age was used throughout all analyses, rather than chronological age, as it provided better model fit (see App.~\ref{app:chron_vs_hear}).

 \subsection{Does embedding distance index concurrent vocabulary size?}

We next examined the relationship between embedding distance and children's vocabulary where we hypothesized that as CHD–FEM embedding distance decreased, vocabulary size would increase, which was consistent across all vocabulary measures (Figs~\ref{fig:model_fits}b–d): embedding distance significantly predicted MB-CDI ($\beta=-0.35$, $p<.001$), PPVT-4 ($\beta=-0.29$, $p<.001$), and EVT-2 ($\beta=-0.24$, $p<.001$), and  scores (Table~\ref{tab:ci_stats}). Model comparisons showed improved fit over baseline when including distance for MB-CDI (e.g., $\Delta$AIC $=-10.28$, $p<.001$), with distance explaining additional variance for all outcomes ($3.51$–$9.47\%$, $p<.001$). Although embedding distance remained a significant predictor of PPVT-4 and EVT-2 scores and explained additional variance, it did not significantly improve model fit. This may indicate that embedding distance captures developmental variance related to receptive and expressive vocabulary, but partially overlaps with age and acoustic covariates, limiting its unique contribution to model fit.

\subsection{Does embedding distance index concurrent consonant articulation skill?}

Our third analysis evaluated whether embedding distance was related to children's consonant articulation ability, measured using the GFTA-2. We expected that smaller CHD–FEM embedding distances would correspond to more accurate consonant articulation, which was confirmed ($\beta=-0.41$, $p<.001$; Table~\ref{tab:ci_stats}; Fig~\ref{fig:model_fits}e), indicating, as expected, that children with more mature, accurate consonant articulation skill are closer in embedding space to global adult speech models. Similar to PPVT-4 and EVT-2 experiments, including distance did not significantly improve overall model fit, but explained substantial additional variance in GFTA-2 scores ($14.33\%$, $p<.001$), suggesting that embedding distance captured articulation-related developmental variation that partially overlapped with hearing age, $f_0$, and voc. length.

\subsection{Are the results robust to diarization errors?}

To assess the robustness of our proposed metric to upstream diarization errors, we conducted sensitivity analyses, simulating increasing levels of speaker misattribution (full results in Appendix~\ref{app:sens_analysis_results}). Across outcomes, the standardized embedding-distance coefficient remained directionally stable, with smaller CHD--FEM embedding distance continuing to predict more mature developmental and language outcomes despite progressive attenuation with increasing contamination (App.~\ref{app:sens_analysis_results}  Fig~\ref{fig:beta_contam}). For hearing age and MB-CDI, embedding distance continued to improve model fit relative to the baseline model across most contamination levels, whereas this pattern was not observed for PPVT-4, EVT-2, and GFTA-2, consistent with the original analyses in which distance effects were nonsignificant (App.~\ref{app:sens_analysis_results} Fig~\ref{fig:delta_aic_contam}). The incremental variance remained directionally stable with progressing contamination like the embedding-distance coefficient (App.~\ref{app:sens_analysis_results} Fig~\ref{fig:pct_var_dist_contam}). All outcomes exhibiting initial significance remained relatively robust under moderate contamination, suggesting the observed effects are unlikely to arise solely from systematic LENA speaker-label errors. 

\section{Conclusion}

Child language research has traditionally relied on labor-intensive manual transcription, limiting scalability; consequently, only \textasciitilde103 of the world’s 7,000+ languages are represented in major language acquisition journals \citep{kidd2022langdiversity}. We propose a simple alternative approach: measuring acoustic distance between children and caregivers within the same community directly from speech embeddings. We found that this distance decreases over development---reflecting convergence to adult speech---and is associated with speech-language outcomes. Critically, across models, developmental periods, and measures, embedding distance remained a significant predictor even after controlling for child age,  $f_0$, and vocalization length, suggesting embedding distance captures variance beyond anatomical maturation. We further show that these results are robust to diarization errors through sensitivity analyses, suggesting the results are not artifacts of speaker diarization error. Together, these results suggest a simple and scalable technique to measure children’s language development from their everyday speech patterns.



\section*{Limitations}

We emphasize that embedding distance is an alternative measure of child language intended to complement, rather than replace, clinical analysis and careful characterization of child language patterns by trained professionals. While promising, the measure does not yet capture specific aspects of language development (e.g., morphological productivity), and its diagnostic utility remains unestablished. There is no evidence that it can be used diagnostically. Furthermore, an acoustic measure based on the child’s own speech production will only indirectly index the child’s receptive capabilities (e.g., ability to distinguish between phonological categories). Although we controlled for mean $f_0$ and vocalization length to account for developmental changes in vocal anatomy and utterance duration, embedding distance may still capture additional acoustic or non-linguistic sources of variation that require further investigation. This metric could also be extended to larger cohorts of children with typical hearing and controlled receptive outcomes, including speech perception tasks, to further evaluate its relationship to receptive language development.

\section*{Ethical considerations}

The model employed, HuBERT-BASE, which was pretrained on 960 hours of English \textit{LibriSpeech} audiobook speech \citep{hsu2021hubert, panayotov2015librispeech}; therefore, the learned representations reflect monolingual English pretraining rather than multilingual exposure. The children reported in this work were primarily acquiring American English. It will thus be critical, going forward, to evaluate the performance of the embedding distance metric, derived from models such as HuBERT, in additional languages that are under-represented in the foundation model's training data, as well as how this approach of embedding distance extends to children acquiring different languages, or combinations of languages.  

 \section*{Acknowledgments}

The authors thank the families who participated in this research, as well as Amy Martinez. Additional thanks to Jan Edwards, Ben Munson, and Mary Beckman for generously sharing their data, portions of which were reused for this project; data collection for those data was originally funded by National Institute on Deafness and Other Communication Disorders grant R01DC02932. Additional data collection and compute resources were funded by a Hearing Health Foundation Emerging Research Grant to M.C. 

\bibliography{custom}

\appendix


\section{Chronological age vs. Hearing age}
\label{app:chron_vs_hear}

\subsection{Methods}

We evaluated model fit using hearing age, defined as months since intervention, and chronological age, defined as months old, and compared their relative ability to explain developmental variance using model fit ($\Delta$AIC) and incremental variance explained ($\Delta R^2$). All models included the baseline predictors of $f_0$, vocalization length, and CHD--FEM embedding distance. They additionally included either chronological age or hearing age, but not both. Because the chronological age and hearing age models are not nested, likelihood ratio tests were not appropriate and were therefore not evaluated. Instead, model comparisons focused on $\Delta$AIC, computed as the AIC of the chronological age model minus the AIC of the hearing age model, such that positive values indicate better model fit using hearing age instead of chronological age. We were not interested in the direction or magnitude of the age coefficients themselves, but rather in whether chronological age or hearing age provided a better overall fit to developmental outcomes beyond the mentioned baseline predictors.

\subsection{Results}

Hearing age generally provided a better account of developmental outcomes than chronological age (Table~\ref{tab:age_ci_stats}). Model comparisons favored hearing age for PPVT-4, EVT-2, and GFTA-2, with lower AIC values relative to chronological age ($\Delta$AIC = 4.92, 3.13, and 2.95, respectively; all $p<.001$), indicating improved model fit. Consistent with this, hearing age explained substantially more additional variance than chronological age for PPVT-4 (19.19\% vs. 6.38\%), EVT-2 (11.06\% vs. 1.34\%), and GFTA-2 (11.14\% vs. 2.30\%). In contrast, MB-CDI showed little difference between age measures: model comparison slightly favored chronological age ($\Delta$AIC = 1.59), and both predictors explained comparable additional variance (13.14\% vs. 12.59\%). These results suggest that hearing age more closely tracks later vocabulary and articulation outcomes, whereas early vocabulary measured by the MB-CDI is similarly captured by either age metric, hence our choice for using hearing age in the main results.

\begin{table*}[!t]
  \centering
  \small
  \begin{tabular}{l c c c c}
    \hline
    \textbf{Statistic} & \textbf{MB-CDI} & \textbf{PPVT-4$^\dagger$} & \textbf{EVT-2$^\dagger$} & \textbf{GFTA-2$^\dagger$} \\
    
    \hline
    \multicolumn{5}{l}{\textit{Model comparison (chron. age vs. hearing age)}} \\
    
    \quad $\Delta$AIC            
        & 1.59 
        & 4.92*** 
        & 3.13*** 
        & 2.95*** \\
    
    \hline
    \multicolumn{5}{l}{\textit{Incremental variance explained ($\Delta R^2$; beyond baseline and distance predictors)}} \\
    
    \quad Chronological age (\%)          
        & 13.14*** 
        & 6.38*** 
        & 1.34** 
        & 2.30** \\
    
    \quad Hearing age (\%)       
        & 12.59*** 
        & 19.19*** 
        & 11.06*** 
        & 11.14*** \\
    
    \hline
    \end{tabular}
    \caption{Bootstrap model summaries. Convention follows Section~\ref{sec:results} Table~\ref{tab:ci_stats}.}
    \label{tab:age_ci_stats}
\end{table*}


\section{Sensitivity analysis}
\label{app:sens_analysis}

\subsection{Methods}
\label{app:sens_analysis_methods}

A limitation of the LENA diarization algorithm is potential speaker-label error and misclassification (e.g. child vocalizations labeled as female adult), which may propagate into downstream analyses \citep{cristia2021lenaanalysis, gautheron2026errors}. To assess the robustness of our findings to such errors, we conducted a sensitivity analysis in which we assumed that ($k\%$) of the subsampled key child (CHD) diarized clips were misclassified. To simulate label contamination, we adopted a simplifying assumption that the existing LENA speaker assignments were correct and, at each subsampling iteration, replaced ($k\%$) of the CHD clips with clips uniformly sampled from the female adult (FEM), male adult (MAL), and other child (OCHD) speaker categories. Importantly, FEM clips introduced during contamination were sampled from a disjoint pool and were excluded from the FEM centroid construction to avoid circularity between contamination samples and the adult reference representation. The resulting contaminated sample sets were processed through the identical statistical modeling and confidence-interval estimation pipeline used in the primary analyses to quantify the robustness of the CHD--FEM embedding distance for estimating hearing age and language outcomes under increasing levels of speaker-label noise.

We did not introduce contamination into the FEM centroid itself. Unlike the CHD sample pool, the FEM representation was constructed as an aggregated centroid across recordings, yielding a single reference point expected to be comparatively stable to individual diarization errors. Contamination was restricted to the CHD samples, reflecting the primary source through which speaker-label noise would influence the child--adult distance.

Analysis was performed as a contamination sweep from 0\% to 100\% in 10\% increments.

\subsection{Results}
\label{app:sens_analysis_results}

\begin{figure*}
\centering
\begin{overpic}[width=1\linewidth]{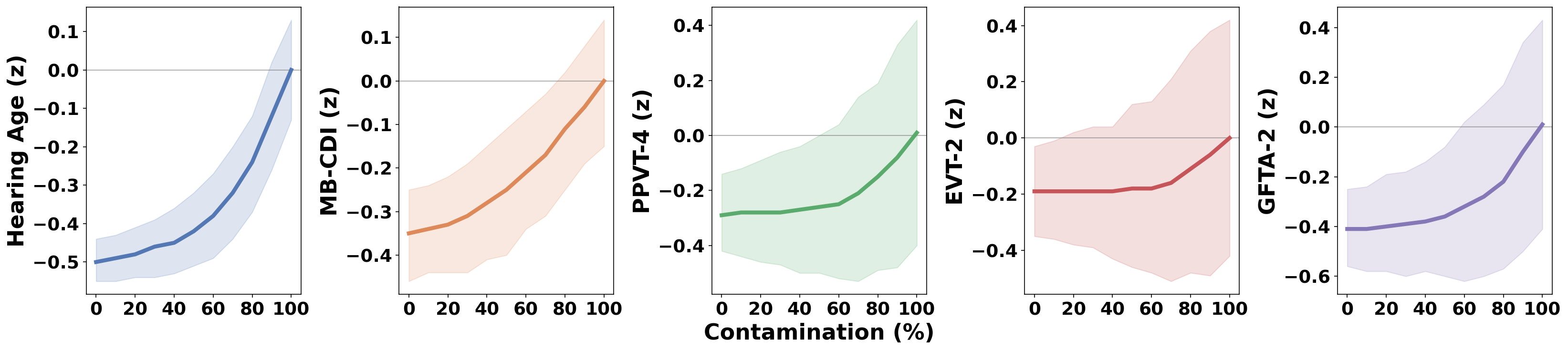}
\end{overpic}
\caption{Standardized embedding-distance coefficient ($\beta$) under simulated speaker-label contamination. Contamination was swept from 0\% to 100\% in 10\% increments. Solid lines show the mean estimated coefficient across bootstrap iterations, and shaded regions denote the corresponding 95\% bootstrap confidence intervals.}
\label{fig:beta_contam}
\end{figure*}

Fig.~\ref{fig:beta_contam} shows how the standardized distance coefficient remained directionally stable across contamination levels. For hearing age and all language outcomes (MB-CDI, PPVT-4, EVT-2, GFTA-2), the estimated effect consistently remained negative, indicating that greater distance from the adult reference distribution continued to predict poorer developmental and language outcomes. Although the magnitude of the effect attenuated progressively with increasing contamination, the direction of the relationship did not reverse, even under severe perturbation levels ($\geq 80\%$ contamination). As expected, statistical significance diminished as contamination increased and the signal-to-noise ratio decreased. 

\begin{figure*}
\centering
\begin{overpic}[width=1\linewidth]{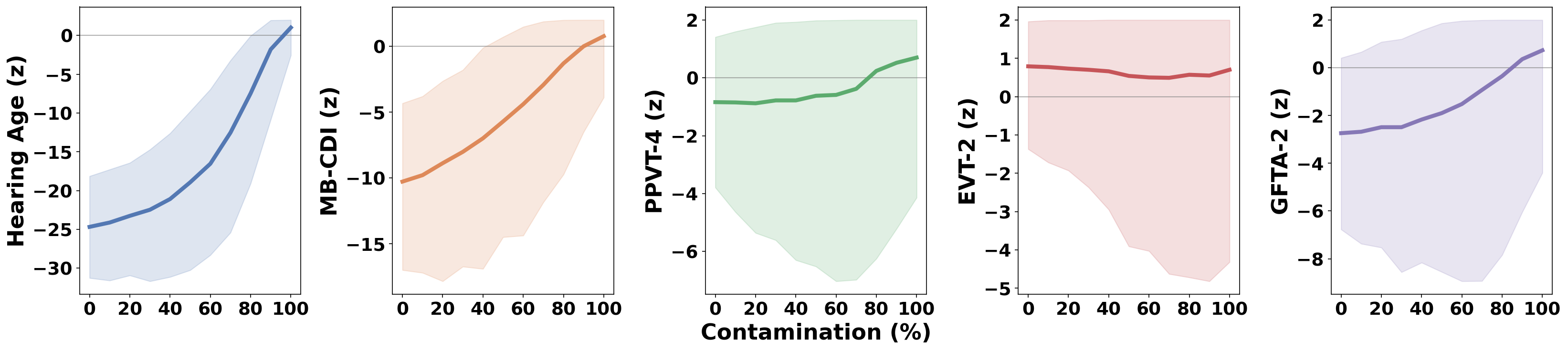}
\end{overpic}
\caption{Model comparison metric ($\Delta$AIC) under simulated speaker-label contamination. Convention follows Figure~\ref{fig:beta_contam}.}
\label{fig:delta_aic_contam}
\end{figure*}

In Fig.~\ref{fig:delta_aic_contam}, across hearing age and MB-CDI, the distance-augmented model (Model 2) consistently outperformed the baseline model (Model 1), as indicated by negative $\Delta$AIC values. Although this advantage diminished with increasing contamination, Model 2 remained preferred across most contamination levels, suggesting that embedding distance provides incremental explanatory value beyond the baseline covariates even under substantial label corruption. In contrast, this pattern was not observed for PPVT-4, EVT-2, and GFTA-2. For these outcomes, the distance effect was already statistically nonsignificant in the original bootstrap analyses, and consequently the contamination sweep did not show a consistent advantage of Model 2 over the baseline model. Thus, the contamination analysis primarily demonstrates robustness for outcomes in which embedding distance exhibited an initial significant association.

\begin{figure*}
\centering
\begin{overpic}[width=1\linewidth]{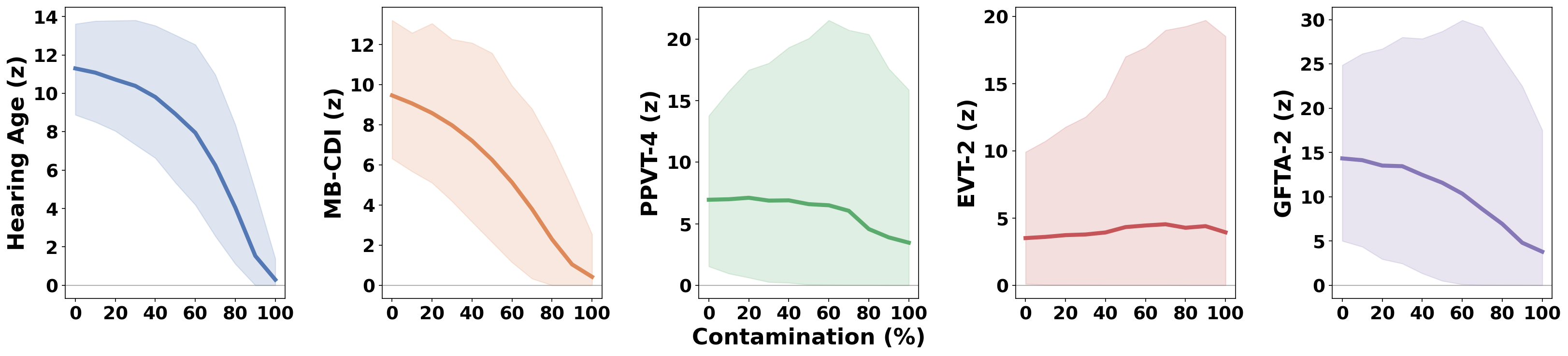}
\end{overpic}
\caption{Incremental variance explained by embedding distance (\% variance explained) under simulated speaker-label contamination. Convention follows Figure~\ref{fig:beta_contam}.}
\label{fig:pct_var_dist_contam}
\end{figure*}

In Fig.~\ref{fig:pct_var_dist_contam}, the incremental variance explained by embedding distance generally decreased with increasing contamination, consistent with progressive degradation of the underlying developmental signal. Across outcomes, this manifested either as a monotonic reduction in the estimated variance explained or as widening confidence intervals that increasingly encompassed the null value. Nevertheless, embedding distance continued to account for a non-trivial proportion of variance across several outcomes, in some cases remaining substantial even under high contamination levels. 

In sum, for outcomes in which the embedding-distance metric exhibited an initial statistically significant association, the estimated effects remained relatively stable under increasing contamination. The progressive attenuation of statistical relationships at higher contamination levels suggests that the observed effects are unlikely to arise solely from systematic speaker-label errors in the LENA diarization pipeline.

\end{document}